\documentclass{article}

\usepackage[final, nonatbib]{neurips_2020}

\usepackage[utf8]{inputenc} 
\usepackage[T1]{fontenc}    
\usepackage{hyperref}       
\usepackage{url}            
\usepackage{booktabs}       
\usepackage{amsfonts}       
\usepackage{nicefrac}       
\usepackage{microtype}      
\usepackage{caption}
\usepackage{subcaption}
\usepackage{graphicx}
\usepackage{float}
\usepackage[parfill]{parskip}
\title{Real-Time Facial Expression Emoji Masking with Convolutional Neural Networks and Homography}

\author{%
  Qinchen Wang$^*$,~ Sixuan Wu$^*$,~ Tingfeng Xia\thanks{Equal contribution} \\\\
  Department of Computer Science \\
  University of Toronto, 40 St. George Street, Toronto, M5S2E4 \\
  \texttt{\{quinn.wang, sixuan.wu, tingfeng.xia\}@mail.utoronto.ca} \\
}

\begin{document}

\maketitle
\begin{abstract}
    Neural network based algorithms has shown success in many applications. In image processing, Convolutional Neural Networks (CNN) can be trained to categorize facial expressions of images of human faces. In this work, we create a system that masks a student's face with a emoji of the respective emotion. Our system consists of three building blocks: face detection using Histogram of Gradients (HoG) and Support Vector Machine (SVM), facial expression categorization using CNN trained on FER2013 dataset, and finally masking the respective emoji back onto the student's face via homography estimation. (\textit{Demo:} \url{https://youtu.be/GCjtXw1y8Pw}) Our results show that this pipeline is deploy-able in real-time, and is usable in educational settings. 
\end{abstract}
\section{Introduction}
During the COVID-19 pandemic, many schools switched to online delivery of course contents. Students are often reluctant to turn on their cameras during classes due to reasons such as being in the bed, did not do makeup, and family members being around. This causes lecturers to lose the very important in-class real-time feedback about the pace of lecturing from attendees’ facial expressions. A typical solution to address this is via associating the attendance grade of the student with whether they turn on their camera during lecture. However, such measure sometimes causes students' adverse emotion, and equity and privacy concerns \cite{barnett, nicardo_khandelwal_weitzman_2020, will_2020}.

In this work, we present our solution: real-time facial expression detection and emoji masking. We propose architecture VGG BA SMALL, which achieves very similar output accuracy for facial expression recognition with a much smaller model size than state-of-art models such as VGG19 \cite{simonyan_zisserman_2014}. We attack the problem of natural looking emoji masking using a non-traditional approach of homography estimation.

\section{Literature Review}
\subsection{Facial Expression Recognition Dataset}
\begin{figure}[]
    \centering
    \includegraphics[width=\textwidth]{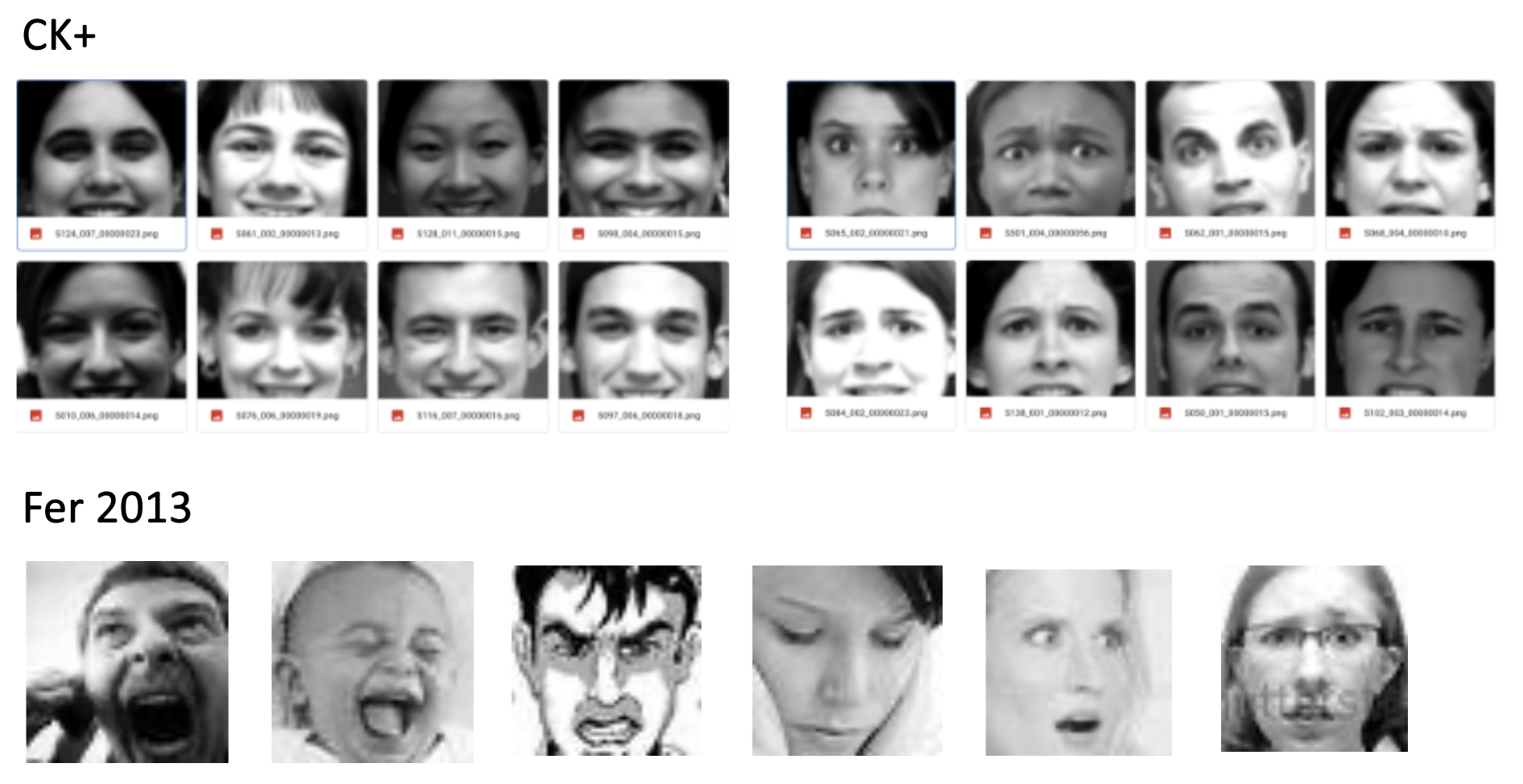}
    \caption{Selected samples from CK+ and FER2013 dataset}
    \label{fig:dataset comparison}
\end{figure}

\begin{figure}[]
    \centering
    \includegraphics[width=\textwidth]{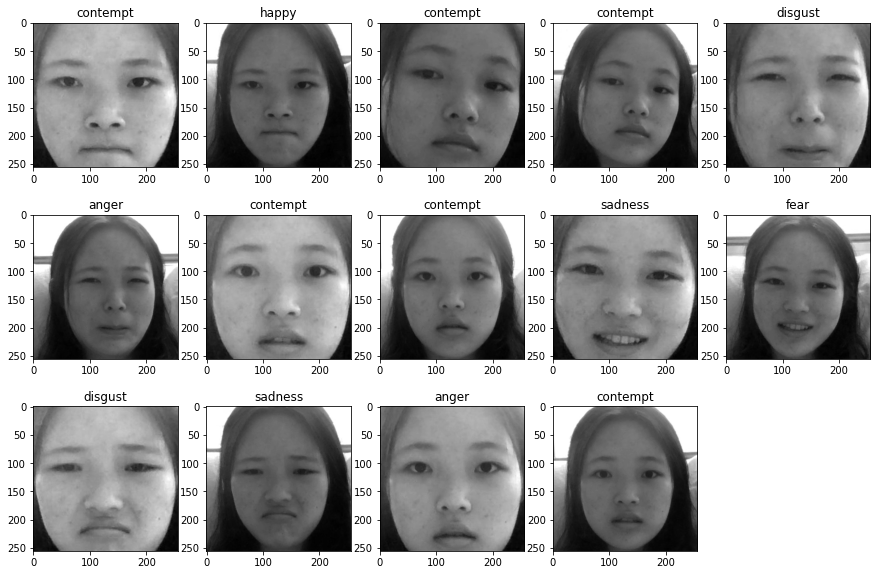}
    \caption{SqueezeNet performance on one of our own faces}
    \label{fig:our own face rec}
\end{figure}

Kulkarni et al. presented a comprehensive review of available facial expression datasets \cite{kulkarni_2009}. Two compelling options, also used extensively in other related literatures, are CK+ and FER2013 \cite{ck+}. Initially, our choice of dataset was CK+ alone, since it is a much smaller dataset and is consist of images with better quality as compared to those in FER2013, as shown by the comparison in Figure \ref{fig:dataset comparison}. The CK+ dataset consists of very well-centered faces. Positions of eyes, nose and mouths are not that different from picture to picture. In contrary, faces in the FER2013 dataset vary in great amount from one to the other. We have variability factors such as position of face, tilting angle of the head, how zoomed in the face is, age of the face, and different expressions in the same emotion class. We expected most students to face directly at the camera, in a position similar to that of images in the CK+ dataset. It therefore seemed sufficient to train with the CK+ dataset for our purposes. It seemed overkill, and a potential source of noise - to include images of babies, cartoons, and ones with watermarks.
However, all models we've experimented perform well (95\%+ accuracy) on CK+, yet they fail dramatically when we try to use the model categorize emotion expressed on pictures of ourselves. One of the most extreme cases is shown by the performance of SqueezeNet \cite{squeezenet}, which yielded a 97\% accuracy on both the training and and validation set segmented from the CK+ dataset, but was unable to show similar levels of performance when it predicts on one of our own faces, as shown in Figure \ref{fig:our own face rec}.

Each expression is classified twice, but sometimes gets classified to different results with a slight move in position. The titles are what’s predicted by the model, and the true label is open to your interpretation, but a lot of these are obviously incorrect. For example, a happy face, cropped into two positions as shown in right most two columns of the second row, are both misclassified, yet one is misclassified as fear and the other one to sadness. We thought of two options to deal with this inconsistency. One is to use the FER2013, we thought maybe more variability in the training set itself would yield a model that predicts well for our own faces. The second is to find a way to crop the face from our own input image, into the same position as those in the CK+ dataset, in real-time. In this work, we will proceed with the first option because we would prefer more generalizability if possible. The model trained on the CK+ dataset is extremely sensitive to face position and we suspect even a very good face cropping algorithm would not be satisfactory. The FER2013 dataset is a much larger data set crawled from the web, with greater variability. We expect a model trained on FER2013 to predict real-world images better. 

\subsection{Facial Expression Recognition Model}
Machine Learning approaches are designed to learn underlying patterns in data without explicit instructions. In the facial expression detection, Convolutional Neural Networks (CNN), which is a type of neural network architecture in machine learning, is used extensively and it has shown state of art results \cite{khan_2020}. Recent works on this topic tend to use an exceptionally large model to improve performance accuracy \cite{Pramerdorfer2016}. State of art models also use an ensemble of multiple CNN to further improve accuracy \cite{goodfellow_2013}.

Our first choice was to use well-known ImageNet oriented architectures, since then we can utilize pretrained models via transfer learning. We experimented with ShuffleNet and SqueezeNet \cite{shufflenet, squeezenet} by training them on the FER2013 dataset. These two specific networks were chosen because they were designed for mobile devices, and thus have a small model size and fast inference speed, much suitable for our real-time criteria. We tested ShuffleNet by training the last fully connected layer with all previous layers' gradient locked. We assessed the SqueezeNet by training it from pretrained weights but with all gradients unlocked. Unfortunately, neither of these two methods work well, giving a final accuracy of only around 20 percent. We observed that weights pretrained from ImageNet do not generalize well to our facial expression classification task, which we suspect is due to the fact that these models learnt a latent space not suitable for the classification of facial expressions. In particular, facial expressions are more fine grained: changing shapes of organs on human face by a slight extent can express something completely different. The ImageNet task, on the other hand, classifies objects, and it might learn more features such as edges and corners. 

Then, we tried to train ShuffleNet, SqueezeNet and Xception net from scratch \cite{xception}. Table \ref{tab:ioa comparisons} shows the best validation accuracy, test accuracy we obtained on FER2013 and model sizes. The Xception model is largest amongst these three tested models, and was designed to outperform the Inception model with a smaller model size \cite{inception, pandit_2020}. The Xception net over-fits dramatically even with dropout layers and $L^2$ regularization. Interestingly, we can observe a trade-off between model size and accuracy: a larger model has more expressive power and can predict better, but it has a slower inference speed. Vice versa.
\begin{table}[]
    \centering
    \begin{tabular}{@{}l|lllll@{}}
        \toprule
        Model                 & ShuffleNet & SqueezeNet  &   Xception Net\\ \midrule
        Valid Acc (\%)        & 50         & 33.60 & 56.45 \\
        Test Acc (\%)         & 50.68         & 33.49 &  56.95\\
        Saved Model Size (MB) & 5          & 3  &   36.43\\ 
        \bottomrule
    \end{tabular}
    \vspace{1em}
    \caption{Validation and testing accuracy of ShuffleNet, SqueezeNet, XceptionNet on FER2013 dataset. Model size indicated in terms of megabytes of the saved PyTorch model. }
    \label{tab:ioa comparisons}
\end{table}

The family of VGG networks are also widely used for the facial expression classification task. We trained VGG11, VGG13, VGG16, and VGG19 \cite{simonyan_zisserman_2014}, on the FER2013 dataset. All models are trained with cross entropy loss, using Stochastic Gradient Descent (SGD) with a momentum of 0.9, a $L^2$ regularization factor of 5e-4, a batch size of 128, and a initial learning rate of 1e-2. Learning rate is decayed by a factor 0.9 every 5 epochs since the 40th epoch until a full stop at 250 epochs. We also enabled gradient clipping in an aide to mitigate the exploding gradient problem \cite{Pascanu2012UnderstandingTE}. Table \ref{tab:vgg comparisons} presents the obtained best validation accuracy, test accuracy, and model size in megabytes for each CNN model. We noticed that despite changes in number of model parameter, the validation and testing accuracy stayed almost the same. This observation inspired us to develop our own modified version of the VGG network, presented later in this work. 

\begin{table}[]
    \centering
    \begin{tabular}{@{}l|lllll@{}}
        \toprule
        Model                 & VGG BA SMALL & VGG11  & VGG13  & VGG16  & VGG19  \\ \midrule
        Valid Acc (\%)        & 70.103         & 69.908 & 70.549 & 70.995 & 70.549 \\
        Test Acc (\%)         & 71.496         & 70.632 & 72.388 & 71.719 & 71.970 \\
        Saved Model Size (MB) & 23.98          & 35.25  & 35.96  & 58.24  & 76.52  \\ 
        \bottomrule
    \end{tabular}
    \vspace{1em}
    \caption{Validation and testing accuracy of VGG networks on FER2013 dataset. Model size indicated in terms of megabytes of the saved PyTorch model. }
    \label{tab:vgg comparisons}
\end{table}

\subsection{Head Position and Pose estimation}
Many approaches exist for head pose estimation and position detection. The most prominent ones are Haar Cascade classifiers, deep learning based models and Histogram of Gradients (HoG) based detectors \cite{haar_cascade, dalal_triggs_2005, murphy2009}. \cite{gupta_2018} presented an comparison between methods. We summarize the pros and cons of each method as follows. 

The Haar Cascade detector has a lightweight and simple architecture. It is computable on CPU and can detect human faces of different ratios. However, this detector tends to misclassify many areas as human face where in fact no human face is present. Also, this model is not robust in presence of occlusions. 

Deep learning models, such as the one developed by Liu et al. and provided in OpenCV is much more robust against occlusions \cite{ssd}. It achieves state of art accuracy and can detect human faces with different ratio and from different orientations. However, its inference speed on CPU is slower than the other two methods that we discuss here. 

The HoG detector has the smallest model size and is the fastest among these tree models on CPU. It performs well on human faces that are generally front facing. It can handle some occlusions, although not as good as deep learning models. It is also worth mentioning that one of its other drawbacks is it performs poorly on faces that are too small (smaller than $80\texttt{px} \times 80\texttt{px}$). However, this will not be our concern since in a web camera, human faces are generally big. 

\subsection{Related Work}
Past work to mask emoji on people's face are generally low on emotion classification accuracy, yet with a slow inference speed and non-natural looking masking. Duncan et al. presented a solution using a custom VGG\_S network trained on their own home-brewed dataset for facial expression detection \cite{duncan_2016}. They achieved a final test accuracy of 57\%. But since this accuracy is reported on their own dataset, we cannot compare the performance between our model and theirs. Moreover, their result masks emoji onto human's faces by simple super-position, without head pose taken into account. The final model presented in that work takes approximately 0.4 second to compute one frame, which means it is only capable of computing 2.5 frames per second in real-time. In this work, we aim to achieve a natural looking emoji masking with high emotion classification accuracy while maintaining a fast inference speed. 

\section{Approach}
\subsection{Facial Expression Recognition}
\begin{figure}[]
    \centering
    \includegraphics[width=\textwidth]{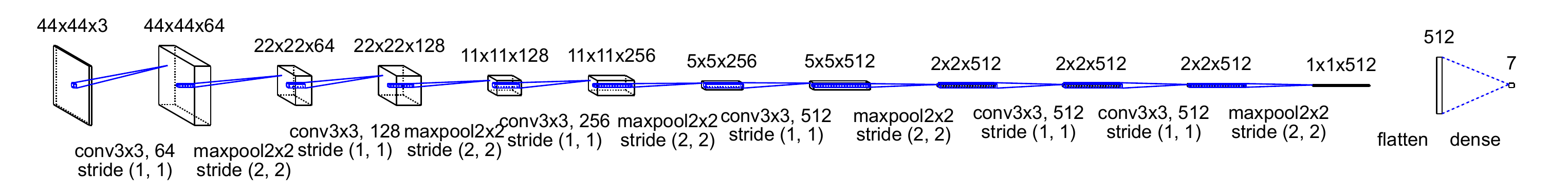}
    \caption{Visualization of VGG BA SMALL architecture, showing only parts that contain trainable parameters. Omitted layers are described: (1) each convolution layer is followed by 2D Batch Normalization and then ReLU non-linearity. (2) Immediately before the dense layer, there is also a 2D average pooling layer. Our code is publicly available at \cite{code}.}
    \label{fig:architecture vgg ba small}
\end{figure}

As shown in Table \ref{tab:vgg comparisons}, despite changes in number of model parameters, the validation and testing accuracy stayed mostly the same. This motivated us to develop the VGG BA SMALL architecture, which is obtained by deleting two convolution layers from VGG11. Refer to our code base for implementation \cite{code}. Figure \ref{fig:architecture vgg ba small} illustrates our architecture. 

\begin{figure}[]
    \centering
    \includegraphics[width=\textwidth]{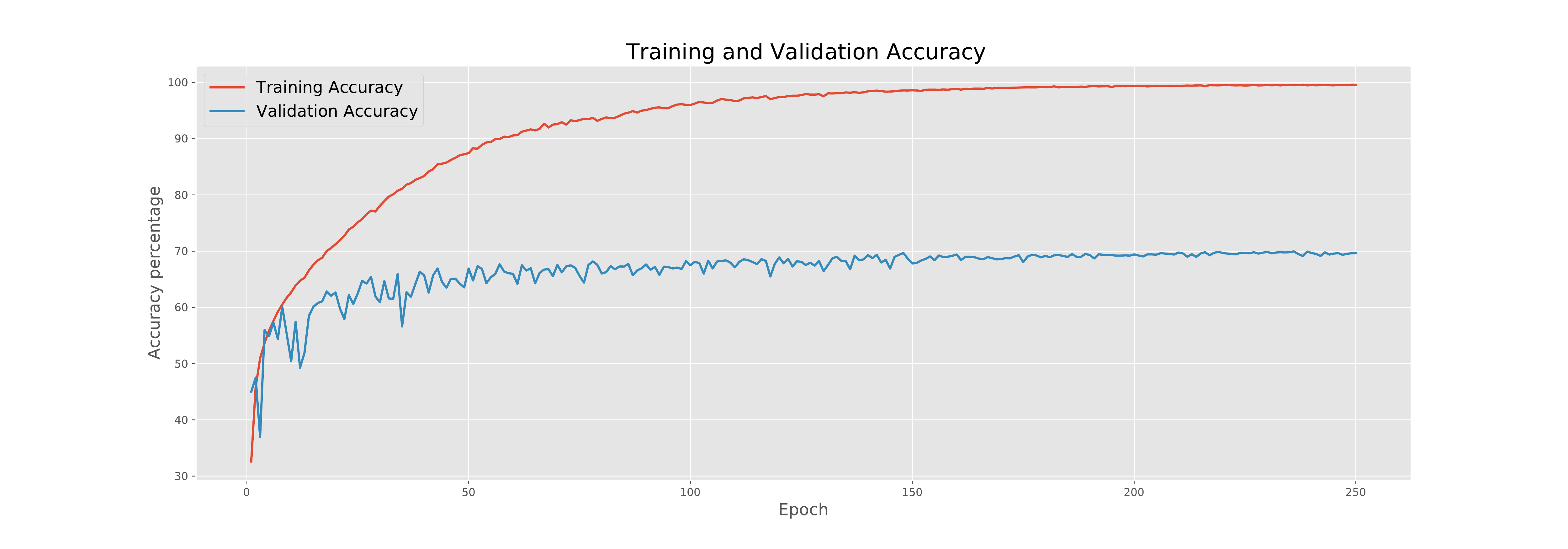}
    \includegraphics[width=\textwidth]{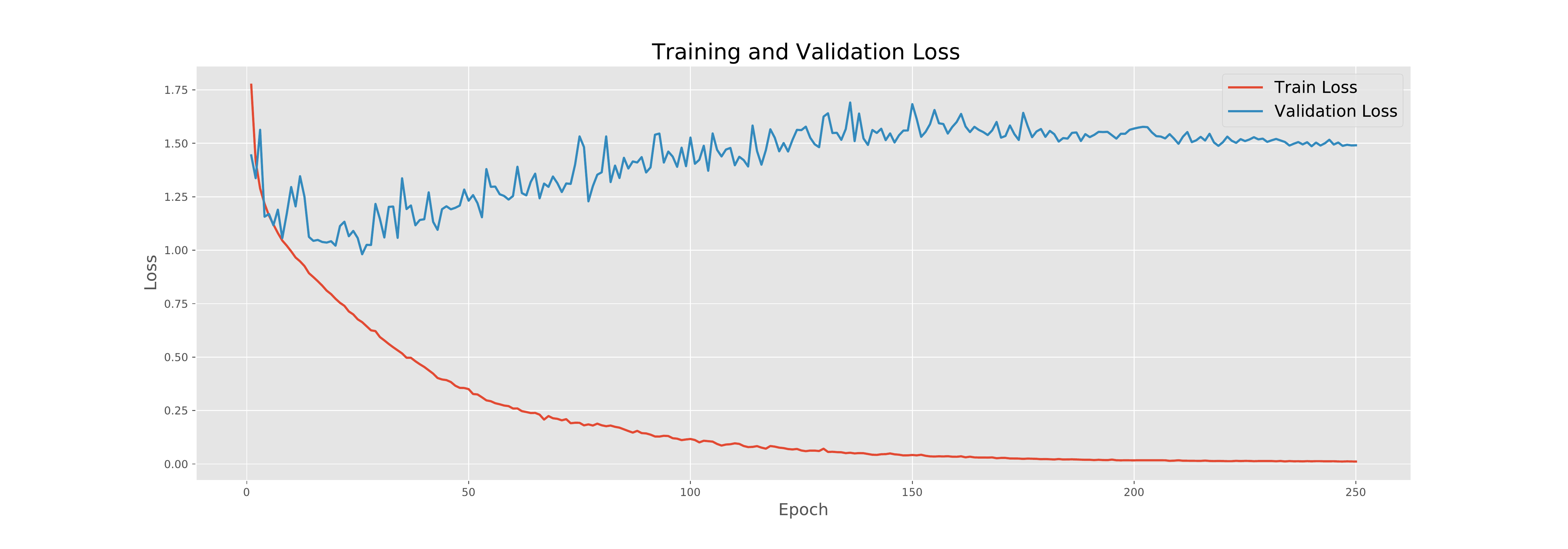}
    \caption{Training and validation accuracy and loss curves for VGG BA SMALL architecture. Loss reported is averaged per-sample loss in each batch.}
    \label{fig:vgg ba small curves}
\end{figure}

\begin{figure}[]
    \centering
    \includegraphics[width=0.8\textwidth]{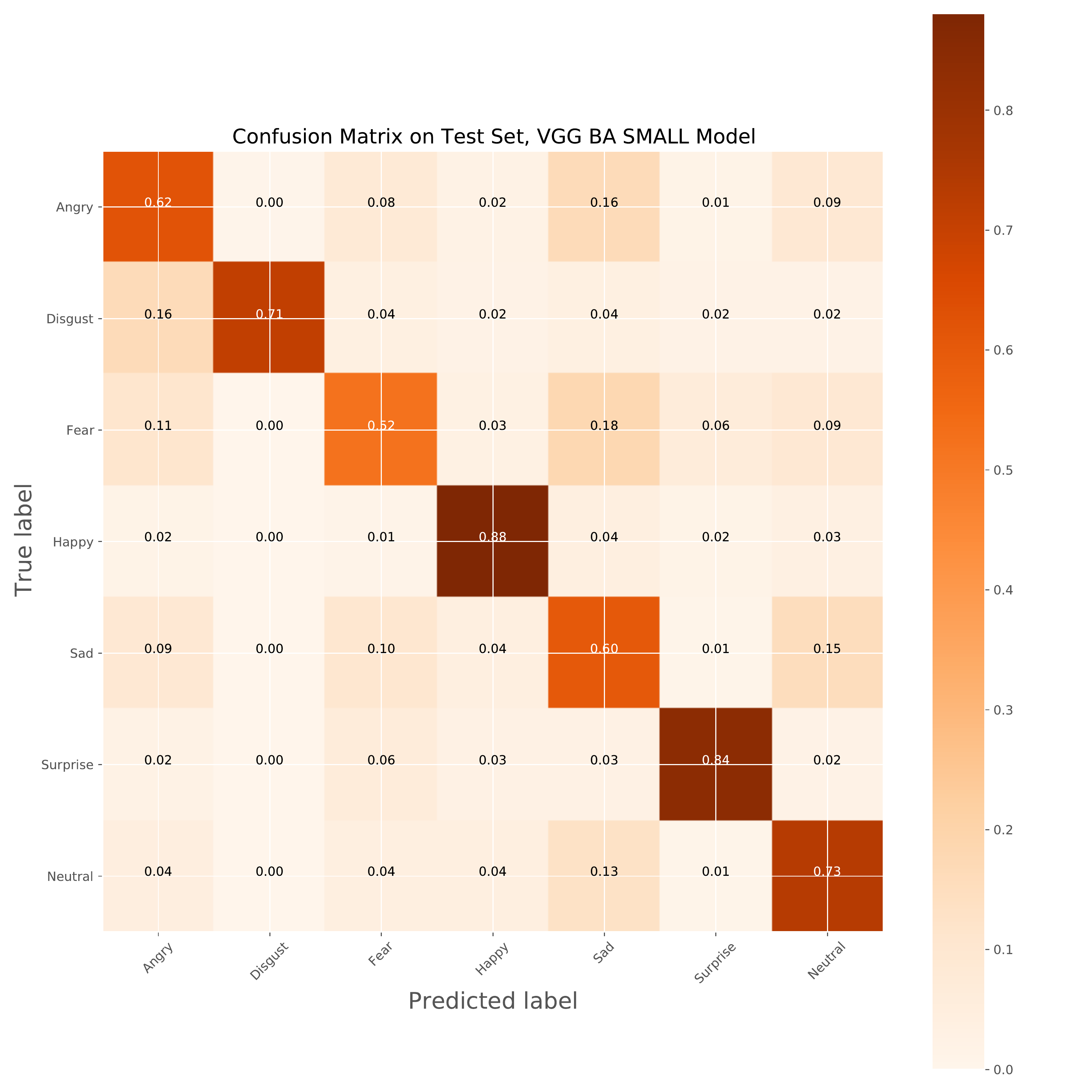}
    \caption{Confusion matrix of VGG BA SMALL final model on test set. }
    \label{fig:vgg ba small confusion}
\end{figure}

We fine tuned our VGG BA SMALL model by performing a grid search on combinations of different batch size, learning rate, learning rate decay factor. Due to limitations in computation power, our search is not too extensive. Our final results indicate that the initial hyper-parameter settings\footnote{Same as our experiments with VGG11, 13, 16, and 19 networks. Trained with cross entropy loss, using SGD with momentum of 0.9, $L^2$ regularization factor of 5e-4, batch size of 128, and a initial learning rate of 1e-2. The learning rate is decayed by a factor of 0.9 every 5 epochs since the 40th epoch until a full stop at 250 epochs. Gradient clipping was enabled during back propagation.} achieves the best accuracy. Results are shown in Table \ref{tab:vgg comparisons}. Figures \ref{fig:vgg ba small curves} and \ref{fig:vgg ba small confusion} show the training curves and test set confusion matrix respectively. The confusion matrix indicates that our model is generally strong in all categories, with possible weakness in sadness, fear and angry classes. Our best result achieves the 70\% classification accuracy tier with just a size of 23.98 Megabytes, much smaller than the established VGG models we mentioned earlier. It is also interesting to notice that the testing accuracies are consistently higher than validation accuracies in our experiments. We suspect this is due to randomness, i.e. random splitting of the dataset yielded a test set that resembles the training set slightly more than the validation set.

\subsection{Head Position and Pose Estimation and Position Detection}
Our target use case is virtual conference platforms, where faces will be mostly front facing and occlusions might happen due to low light or bad web cam quality. Hence, we chose the HoG + SVM approach because it is the fastest among these three approaches on CPU, suitable for front face detection and it is tolerant to occlusions. Our implementation uses the ``\texttt{shape\_predictor\_68\_face\_landmarks}'' key point HoG + SVM detector from Dlib. 

\subsection{Emoji Masking\label{Emoji Masking}}
We perform natural masking of emoji via homography estimation. To do so, we first label key points, such as chin and eye corners, locations on the emoji. Our labels follow the proportion rules exhibited in drawing, as discussed in \cite{tds, fussell}. In total, there are six key points: left eye outer corner, right eye outer corner, mouth left corner, mouth right corner, philtrum, and chin. At the first glance, philtrum seems like a odd choice and it is compelling to mark nose tip as one of the key points, since it is at the centre of a human's face. However, doing so breaks the planer object assumption of homography estimation. 

\begin{figure}
    \centering
    \begin{subfigure}[b]{0.49\textwidth}
        \centering
        \includegraphics[width=0.89\textwidth]{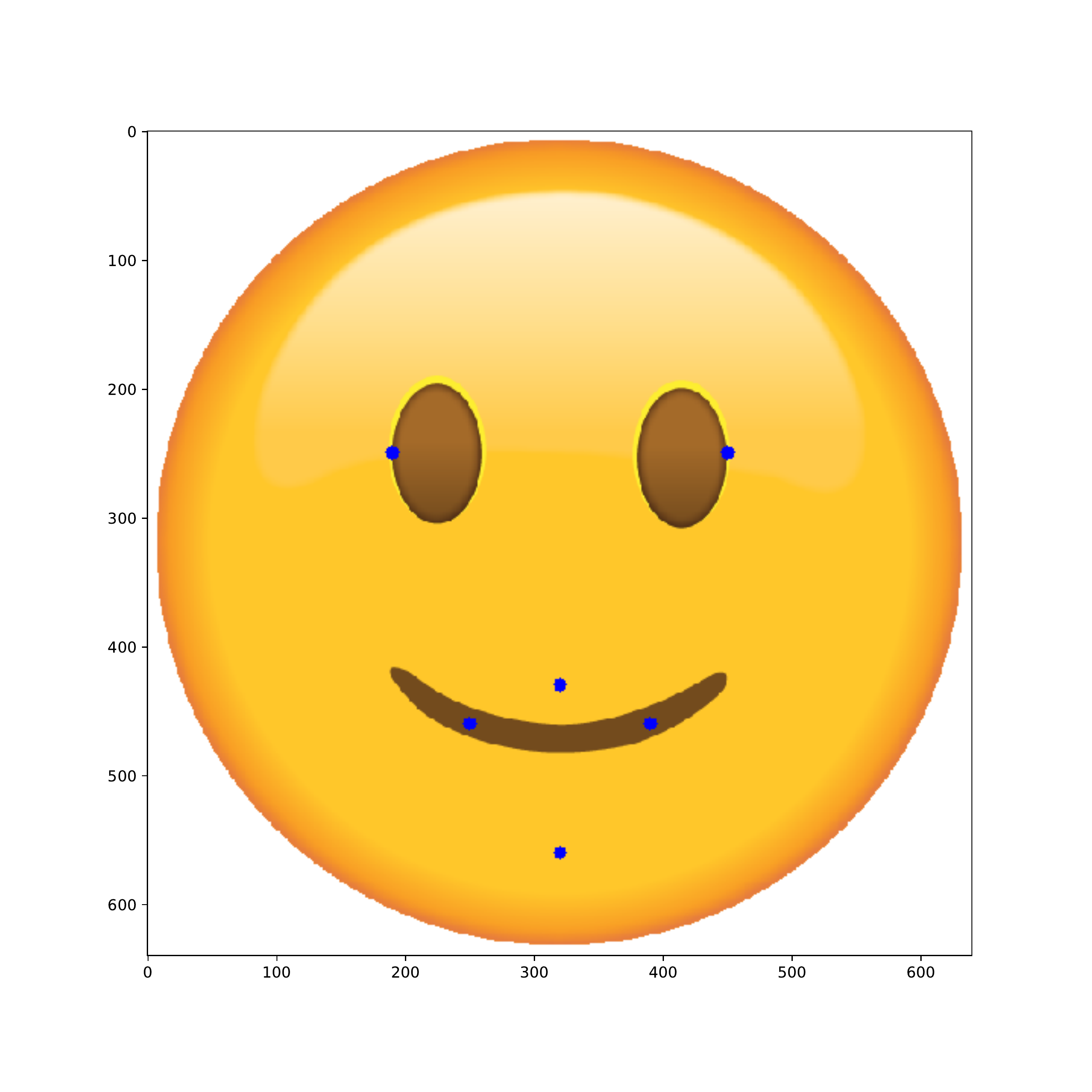}
        \caption{Emoji marked with key points. Coordinates calculated based on human face ratios as used in drawing. See discussion in Section \ref{Emoji Masking} for reference. This emoji is chosen to correspond to the ``neutral'' label in this work.}
        \label{fig:emoji marked}
    \end{subfigure}
    \hfill%
    \begin{subfigure}[b]{0.48\textwidth}
        \centering
        \includegraphics[width=\textwidth]{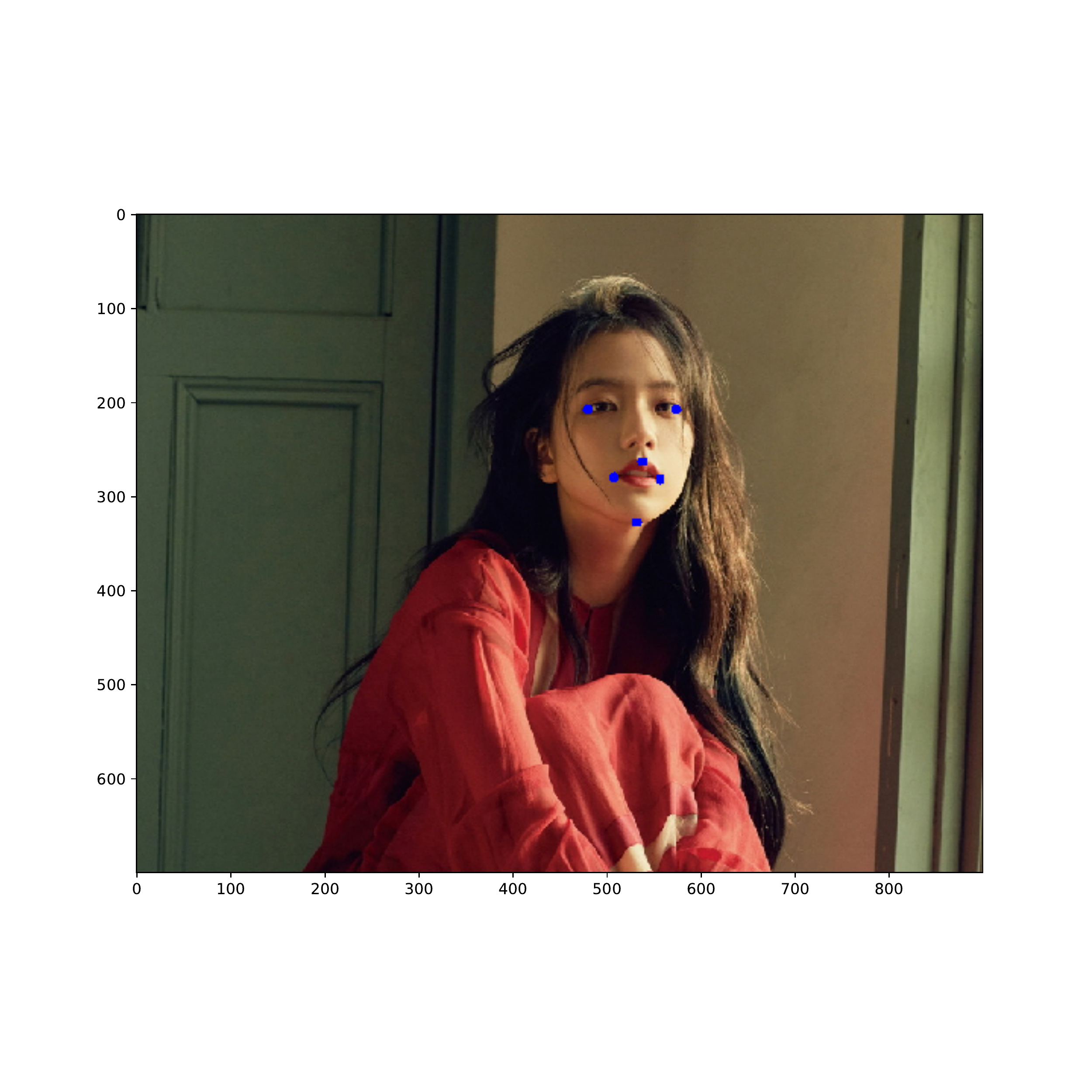}
        \caption{Human face example marked with key points. Coordinates estimated by Dlib HoG + SVM 68 facial landmark model. Marking based on portrait of Kim Jisoo \cite{jisoo}.}
        \label{fig:face marked}
        \vspace{0.8em}
    \end{subfigure}
    \caption{Emoji and human face with marked key points: outer corners of both eyes, philtrum, outer corners of mouth, and chin. }
    \label{fig: human face and emoji}
\end{figure}

We then estimate the homography transformation by using coordinates of key points on the emoji as source and coordinates of key points on human's face as destination. Figures \ref{fig:emoji marked} and \ref{fig:face marked} shows the key points on an example picture and an emoji. The resulting masked image can then be simply obtained by superposing the original image, and the homography warped version of the emoji. 
\subsection{Pipeline}
\begin{figure}
    \centering
    \includegraphics[width=0.7\textwidth]{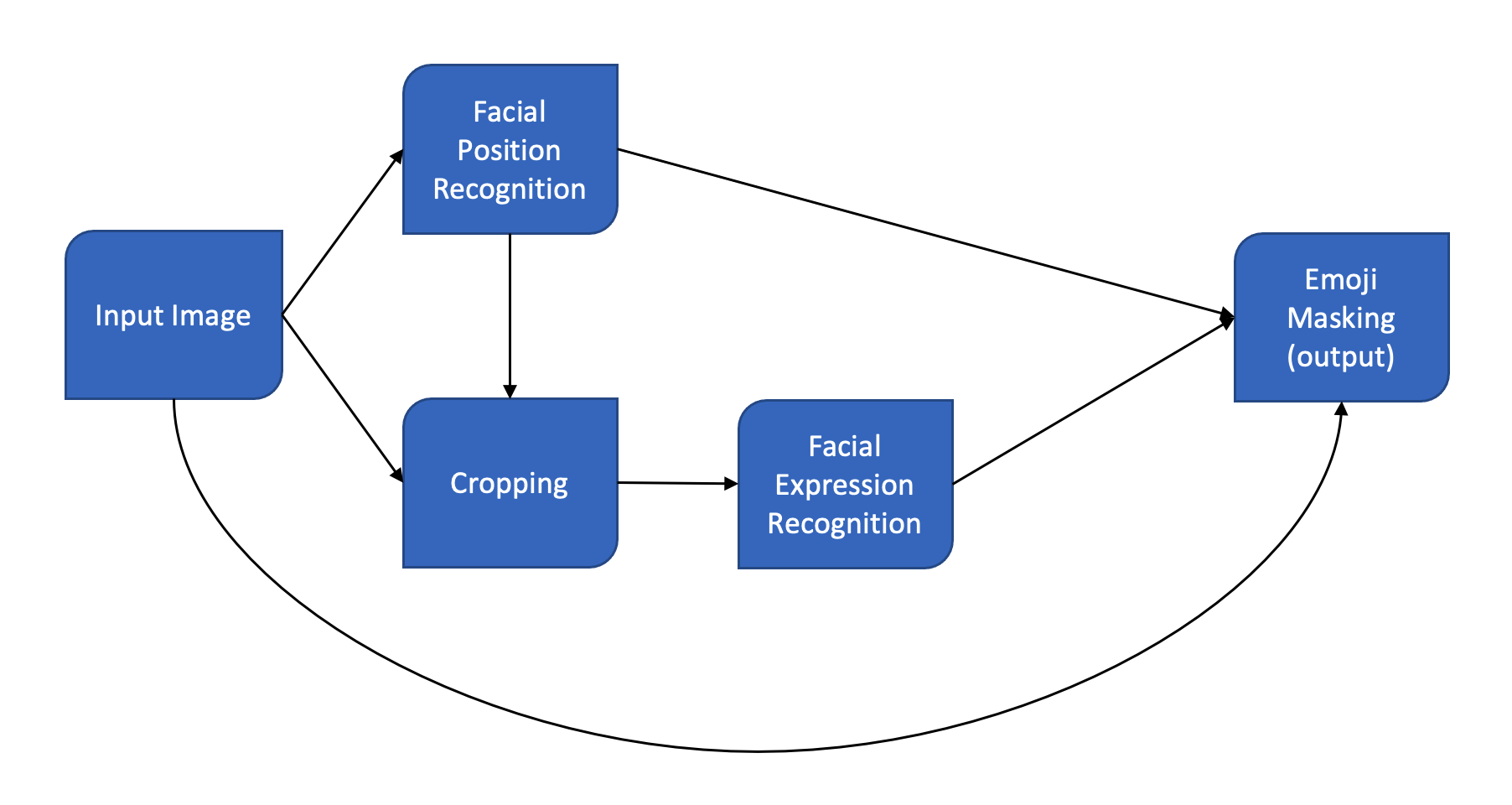}
    \caption{Processing pipeline}
    \label{fig:pipeline}
\end{figure}

In this section, we present our processing pipeline, as shown in Figure \ref{fig:pipeline}. The pipeline can be modularized into three sections:
\paragraph{Face Detection} INPUT: one frame of video image. OUTPUT: section of face cropped out from image. Method: we use the Dlib library to get the location of the frontal face, cropped as a rectangle.
\paragraph{Expression Detection} INPUT: section of the face cropped out from image. OUTPUT: emotion classification of the face from the model. Method: we transform the input image into a $44 \times 44$ gray-scale image and feed it into the VGG BA SMALL model.
\paragraph{Emoji Masking} INPUT: frame of the video and classification from the model. OUTPUT: frame of the video with emoji masked onto the face. Method: with the output from the VGG BA SMALL model, we find the corresponding emoji. Each of the 7 emojis has been marked locations of outer eye corners, corners of lips, chin and the philtrum. These location marks act as key points that would later match with corresponding facial landmark locations recognized by Dlib 68. Using these 6 matching key points, we calculate a homography matrix and use it to warp and mask the emoji onto the original image.

\section{Results}
We prepared a demo video, showcasing the real-time detection and masking (available at  \url{https://youtu.be/GCjtXw1y8Pw}). The first part of the video aims to demonstrate our natural masking algorithm using homography estimation. During the second part of the demo, we experimented with switching of facial expressions. We notice that our detection algorithm works particularly well with neutral facial expression. In general, the model performs well in cases where the facial expression is at least slightly exaggerated. However, during the changing from facial expression A to B, the model can show erratic behavior: predicting a third class C for a very few amount of frames in between. 

\begin{figure}[]
    \centering
    \includegraphics[width=\textwidth]{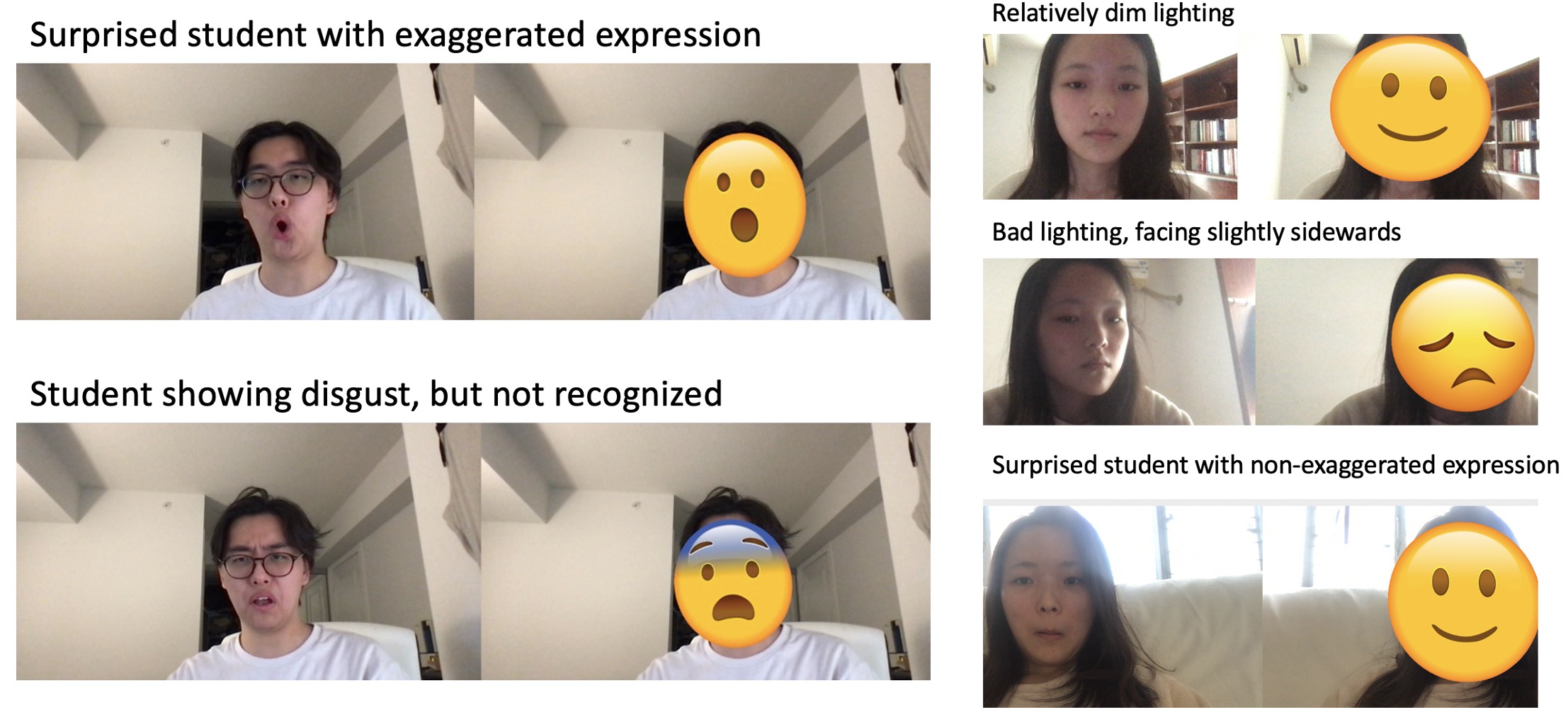}
    \caption{Examples of good performance and limitations}
    \label{fig:good and bads}
\end{figure}

We tested for some more extreme cases and summarized when the model performed well and its limitations. With respect to lighting conditions, the model works well even in a dim environment. However, if the lighting becomes extremely bad, the performance drops. With respect to angles of face, the model works well if the student is looking down at the camera. However, if the camera was above the student, the model is susceptible to predicting emotions such as ``anger'' and ``sadness'' even when the emotion is in fact neutral. The model becomes sometimes erratic for students facing the camera with their sides (left or right) - a wrong prediction would jump out amongst consecutive correct predictions, and as a result you might see the emoji mask change even when you are not moving. Also mentioned in the future work section, this is why we believe integrating a majority vote system - for example, masking only the emoji corresponding to the class predicted most in the last 5 predictions - would improve the model robustness to these erratic behaviours. With respect to the level of exaggeration needed, the model works well if the student expresses emotion with an obvious change in facial features. However, the limitation here is that many emotions are subtle. For example, a slightly confused student may be recognized as having a neutral expression. Furthermore, many students don't show emotions through exaggerated expressions and the model compresses images to only 44x44, further eliminating subtle emotion features in the frame. This leads us to what we mention in our future work section - analysing body language as well as facial features. With respect to specific emotions, we noticed that the model is very good at recognizing neutral expressions, but rarely predicts an emotion as "disgust", even though the student is clearly showing an exaggerated expression showing disgust. We suspect this likely due to an imbalanced class issue in the FER2013 dataset, where after splitting the train, validation and test set, there are only 437 examples in the training set, where any of the other classes has at least 3000 examples. Illustrations of some cases mentioned in this paragraph is shown in Figure \ref{fig:good and bads}.

To assess the model performance, we timed the processing time for an array of 373 frames using solely CPU for inference.\footnote{Experiment conducted on machine with Intel(R) Core(TM) i7-9750H CPU, input is at 720P.} Excluding the time to load the models from files, the model took in total approximately 28.21 seconds to process and generate the masked images for all 373 frames. This results in a average processing time of approximately 0.07563 seconds per frame. Thus, our model is capable of real-time emoji masking at a rate of 13 frames per second. 
\section{Conclusion}
The goal of this project was to implement a real-time facial expression detection algorithm. We use natural masking of the emoji to present the results in an aide to encourage students to turn on their camera during lectures. Using our own VGG BA SMALL architecture for emotion detection with face landmark detector provided by Dlib, we successfully implemented an application wherein a respective emoji is masked naturally via homography back onto human's face. Although we achieved a successful implementation, significant improvements can be made to further polish this project. We discuss this by presenting some of our ideas in Section \ref{sec: future work} Future Work. 

\section{Future Work\label{sec: future work}}
Our VGG BA SMALL network was trained on the FER2013 dataset, which is an imbalanced dataset in terms of ethnicity of people in the picture samples. For example, samples of Asian faces are rare. Future work to address this could blend in other data sets, such as JAFFE \cite{jaffe}, to create a balanced dataset. 

In a real-time video application where consecutive frames could be obtained, future researchers can also integrate the idea of majority vote to improve robustness. We expect that the majority vote will not affect the prediction accuracies, although it is likely to improve the robustness of the pipeline in the case of processing videos. Another possibility is to use recurrent neural networks to capture the history to aid the stabilization of current frame prediction. 

Future work could also animate the emoji to make the masking look even more natural. This includes, for example, stretching the emoji face in horizontal and vertical directions to accommodate different face ratios and hyperbole expressions.

Although our pipeline is usable in real-time scenarios, with a 13 frames per second performance, we expect improvements to be made by utilizing concurrency. For example, when consecutive frames are fed into the model, the face position detector can start its work prior to the finishing of computation of the previous frame.

Our work uses solely facial expression to detect emotion, while in fact human emotions can also be expressed via body language. In particular, our solution does not work well for people with non-expressive facial expressions. Future researchers could investigate approaches to integrate facial expression information and body language detection to build a more robust model with greater applicability. For our project setting, an example of body language could be the tightness in shoulders and necks.

\section{Authors' Contributions}
Our group did everything more or less together, but here is a rough breakdown of each member's major contribution. 
\begin{itemize}
    \item \textbf{Qinchen}: facial expression data set comparison, SqueezeNet prototyping (including transfer learning and training with unlocked gradients), emoji key point marking, code base management. 
    \item \textbf{Sixuan}: ShuffleNet prototyping (including transfer learning and training from scratch), head pose detection, emoji key point marking, merging pipeline, recording demo. 
    \item \textbf{Tingfeng}: Xception and VGG nets prototyping (training from scratch), head pose estimation, homography masking, emoji key point marking, write up management. 
\end{itemize}

\bibliographystyle{unsrt}
\bibliography{bib.bib}
\end{document}